\documentclass[conference]{IEEEtran}
\usepackage{times}

\usepackage{multicol}
\usepackage[bookmarks=true]{hyperref}

\usepackage{times}
\usepackage{latexsym}
\usepackage{graphicx}
\usepackage{amsmath}
\usepackage{amssymb}
\usepackage{amsfonts}
\usepackage{booktabs}
\usepackage{bbm}
\usepackage{multicol}
\usepackage{algorithm}
\usepackage[noend]{algpseudocode}
\usepackage{hyperref}
\usepackage[dvipsnames]{xcolor}
\usepackage{float}
\usepackage{wrapfig}
\usepackage[export]{adjustbox}
\usepackage[normalem]{ulem}
\usepackage[dvipsnames]{xcolor}

\usepackage[backend=biber,
            hyperref=true,
            url=false,
            isbn=false,
            doi=false,
            backref=false,
            style=ieee,
            citestyle=numeric-comp,
            sorting=nyt,
            maxcitenames=99,
            block=none]{biblatex}

\usepackage[font=footnotesize]{caption}
\newcommand{\code}[1]{\textcolor{RoyalBlue}{$\rhd$ \emph{#1}}}

\DeclareUnicodeCharacter{0301}{\'{i}}

\begin{document}

\title{Robot Learning on the Job: Human-in-the-Loop\\Autonomy and Learning During Deployment}

\author{\authorblockN{
Huihan Liu$^{*}$, Soroush Nasiriany, Lance Zhang, Zhiyao Bao, Yuke Zhu
}
\authorblockA{
The University of Texas at Austin\\
}
}

\maketitle

\renewcommand{\thefootnote}{\fnsymbol{footnote}}
\footnotetext[1]{Correspondence: \url{huihanl@utexas.edu}}
\renewcommand*{\thefootnote}{\arabic{footnote}}

\begin{abstract}
With the rapid growth of computing powers and recent advances in deep learning, we have witnessed impressive demonstrations of novel robot capabilities in research settings. Nonetheless, these learning systems exhibit brittle generalization and require excessive training data for practical tasks. To harness the capabilities of state-of-the-art robot learning models while embracing their imperfections, we present Sirius, a principled framework for humans and robots to collaborate through a division of work. In this framework, partially autonomous robots are tasked with handling a major portion of decision-making where they work reliably; meanwhile, human operators monitor the process and intervene in challenging situations. Such a human-robot team ensures safe deployments in complex tasks. Further, we introduce a new learning algorithm to improve the policy's performance on the data collected from the task executions. The core idea is re-weighing training samples with approximated human trust and optimizing the policies with weighted behavioral cloning. We evaluate Sirius in simulation and on real hardware, showing that Sirius consistently outperforms baselines over a collection of contact-rich manipulation tasks, achieving an 8\% boost in simulation and 27\% on real hardware than the state-of-the-art methods in policy success rate, with twice faster convergence and 85\% memory size reduction. Videos and more details are available at \url{https://ut-austin-rpl.github.io/sirius/}
\end{abstract}

\IEEEpeerreviewmaketitle

\section{Introduction}

Recent years have witnessed great strides in deep learning techniques for robotics. In contrast to the traditional form of robot automation, which heavily relies on human engineering, these data-driven approaches show great promise in building robot autonomy that is difficult to design manually. While learning-powered robotics systems have achieved impressive demonstrations in research settings~\cite{andrychowicz2020learning,kalashnikov2018qt,lee2020learning}, the state-of-the-art robot learning algorithms still fall short of generalization and robustness for widespread deployment in real-world tasks. The dichotomy between rapid research progress and the absence of real-world application stems from the lack of performance guarantees in today's learning systems, especially when using black-box neural networks. It remains opaque to the potential practitioners of these learning systems: how often they fail, in what circumstances the failures occur, and how they can be continually enhanced to address them.

To harness the power of modern robot learning algorithms while embracing their imperfections, a burgeoning body of research has investigated new mechanisms to enable effective human-robot collaborations.
Specifically, {\em shared autonomy} methods~\cite{Javdani2015SharedAV,Reddy2018SharedAV} aim at combining human input and semi-autonomous robot control to achieve a common task goal. These methods typically use a pre-built robot controller rather than seeking to improve robot autonomy over time. Meanwhile, recent advances in \textit{interactive imitation learning}~\cite{kelly2019hg,mandlekar2020humanintheloop,ross2011reduction,celemin2022interactive} have aimed to learn policies from human feedback in the learning loop. Although these learning algorithms can improve the overall efficacy of autonomous policies, these policies still fail to meet the performance requirements for real-world deployment.

\begin{figure}[t]
    \centering
    \includegraphics[width=1.0\linewidth]{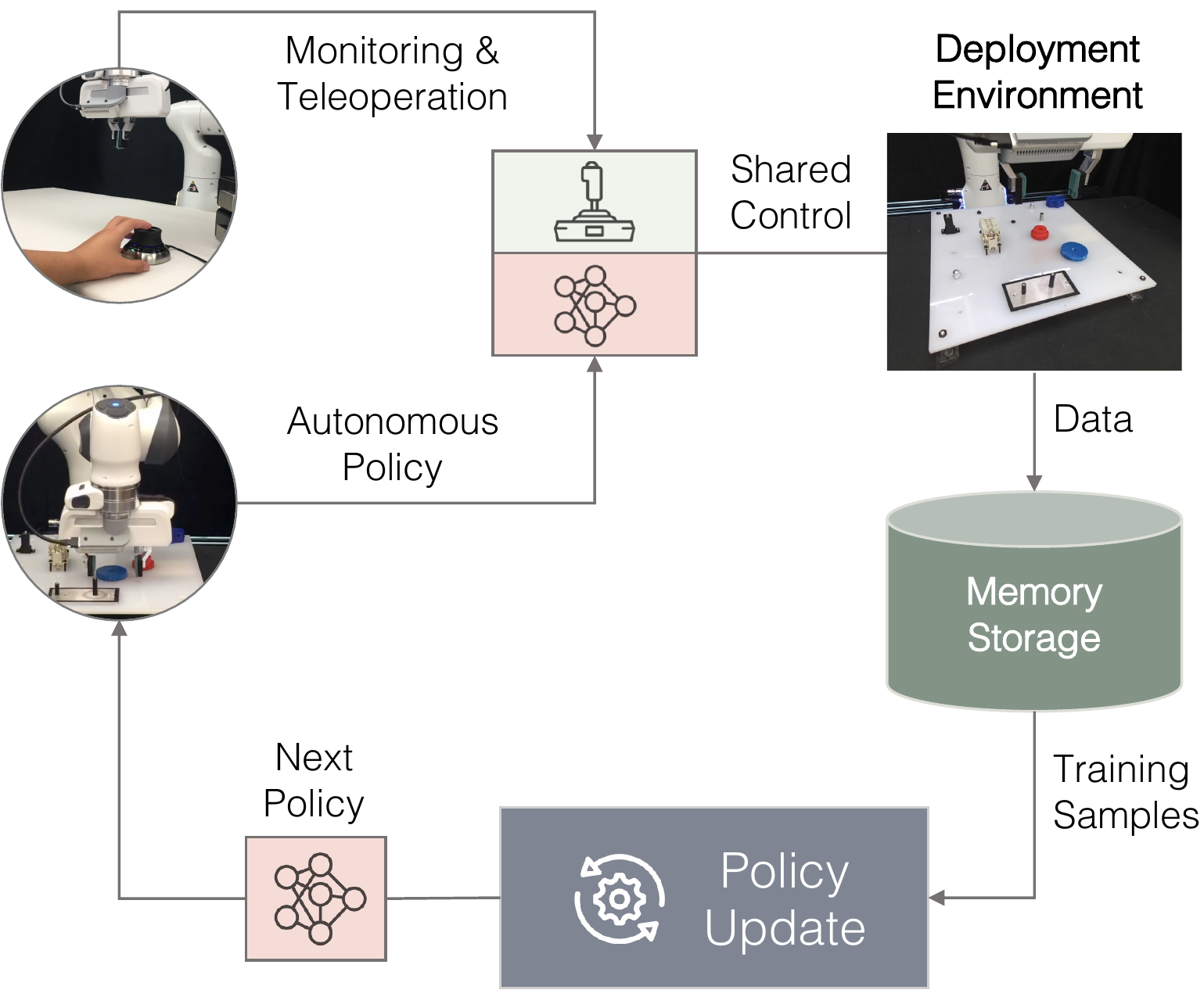}
    \caption{\textbf{Overview of Sirius, our human-in-the-loop learning and deployment framework.} Sirius enables a human and a robot to collaborate on manipulation tasks through shared control. The human monitors the robot's autonomous execution and intervenes to provide corrections through teleoperation. Data from deployments will be used by our algorithm to improve the robot's policy in consecutive rounds of policy learning.}
    \label{fig:pull-figure}
\end{figure}

This work aims at developing a human-in-the-loop learning framework for human-robot collaboration and continual policy learning in deployed environments. We expect our framework to satisfy two key requirements: 1) it ensures task execution to be consistently successful through human-robot teaming, and 2) it allows the learning models to improve continually, such that human workload is reduced as the level of robot autonomy increases. 
To build such a framework, This idea of \textit{robot learning on the job} resembles the Continuous Integration, Continuous Deployment (CI/CD) principles in software engineering~\cite{CICD}. Realizing this idea for learning-based manipulation invites fundamental challenges.

The foremost challenge is developing the infrastructure for human-robot collaborative manipulation. We develop a system that allows a human operator to monitor and intervene the robot's policy execution (see Fig.~\ref{fig:pull-figure}). The human can take over control when necessary and handle challenging situations to ensure safe and reliable task execution. Meanwhile, human interventions implicitly reveal the \textit{task structure} and the level of \textit{human trust} in the robot. As recent work~\cite{kelly2019hg,mandlekar2020humanintheloop, Hoque2021ThriftyDAggerBN} indicates, human interventions inform \textit{when} the human lacks trust in the robot, \textit{where} the risk-sensitive task states are, and \textit{how} to traverse these states. We can thus take advantage of the occurrences of human interventions during deployments as informative signals for policy learning.

The subsequent challenge is updating policies on an ever-growing dataset of shifting distributions. As our framework runs over time, the policy would adapt its behaviors through learning, and the human would adjust their intervention patterns accordingly. Deployment data from human-robot teams can be multimodal and suboptimal. Learning from such deployment data requires us to selectively use them for policy updates. We want the robot to learn from good behaviors to reinforce them and also to recover from mistakes and deal with novel situations. At the same time, we want to prevent the robot from copying bad actions that would lead to failure. Our key insight is that we can assess the importance of varying training data based on human interventions for policy learning.

To this end, we develop a simple yet effective learning algorithm that uses the occurrences of human intervention to re-weigh training data. We consider the robot rollouts right before an intervention as ``low-quality'' (as the human believes the robot is about to fail) and both human demonstrations and interventions as ``high-quality'' for policy training. We label training samples with different weights and train policies on  these samples using weighted behavioral cloning, the state-of-the-art algorithm for imitation learning \cite{sasaki2021behavioral, DBLP:journals/corr/abs-2011-13885, pmlr-v162-xu22l} and offline reinforcement learning \cite{wang2020critic, nair2021awac, Kostrikov2021OfflineRL}. This supervised learning algorithm lends itself to the efficiency and stability of policy optimization on our large-scale and growing dataset.

Furthermore, deploying our system in long-term missions leads to two practical considerations:
1) it incurs a heavy burden of memory storage to store all past experiences over a long duration, and 2) a large number of similar experiences may inundate the small subset of truly valuable data for policy training.
We thus examine different memory management strategies, aiming at adaptively adding and removing data samples from the memory storage of fixed size. Our results show that even with 15\% of the full memory size, we can retain the same level of performance or achieve even better performance than keeping all data, and moreover enables three times faster convergence for rapid model updates between consecutive rounds.

We name our framework Sirius, the star symbolizing our human-robot team with its binary star system. We evaluate Sirius in two simulated and two real-world tasks requiring contact-rich manipulation with precise motor skills. Compared to the state-of-the-art methods of learning from offline data~\cite{nair2021awac, Kostrikov2021OfflineRL,robomimic2021} and interactive imitation learning~\cite{mandlekar2020humanintheloop}, Sirius achieves higher policy performance and reduced human workload. Sirius reports an 8\% boost in policy performance in simulation and 27\% on real hardware over the state-of-the-art methods.
\section{Related Work}
\label{sec:related}

\textbf{\emph{Human-in-the-loop Learning.}} A human-in-the-loop learning agent utilizes interactive human feedback signals to improve its performance~\cite{zhang2019leveraging,Cruz2020ASO,Cui2021UnderstandingTR}.
Human feedback can serve as a rich source of supervision, as humans often have a priori domain information and can interactively guide the agent with respect to its learning progress.
Many forms of human feedback exist, such as interventions~\cite{kelly2019hg, spencer2020wil, mandlekar2020humanintheloop}, preferences~\cite{christiano2017preferences, biyik2022learning, lee2021pebble, Wang2021SkillPL}, rankings~\cite{Brown2019ExtrapolatingBS}, scalar-valued feedback~\cite{macglashan2017interactive, warnell2018deep}, and human gaze~\cite{ijcai2020-689}.
These feedback forms can be integrated into the learning loop through learning techniques such as policy shaping~\cite{knox2009tamer,NIPS2013_e034fb6b} and reward modeling~\cite{daniel2014active,leike2018scalable}, enabling model updates from asynchronous policy iteration loops~\cite{2022correct_me}.

Within the context of robot manipulation, one approach is to incorporate human interventions in imitation learning algorithms~\cite{kelly2019hg, spencer2020wil, mandlekar2020humanintheloop}.
Another approach is to employ deep reinforcement learning algorithms with learned rewards, either from preferences~\cite{lee2021pebble, Wang2021SkillPL} or reward sketching~\cite{cabi2019scaling}. While these methods have demonstrated higher performance compared to those without humans in the loop, they require a large amount of supervision from humans and also fail to incorporate human control feedback in deployment into the learning loop again to improve model performance. In contrast, we specifically consider the above scenarios which are critical to real-world robotic systems.

\textbf{\emph{{Shared Autonomy.}}} Human-robot collaborative control is often necessary for real-world tasks when we do not have full robot autonomy while full human teleoperation control is burdensome. In shared autonomy \cite{dragan2013policy, Javdani2015SharedAV, gopinath2017human, Reddy2018SharedAV}, the control of a system is shared by a human and a robot to accomplish a common goal \cite{Tan2021InterventionAS}. The existing literature on shared autonomy focuses on efficient collaborative control from human intent prediction \cite{dragan2008formalizing, Muelling2015AutonomyIT, 7140066}. However, they do not attempt to learn from human intervention feedback, so there is no policy improvement. We examine a context similar to that of shared autonomy where human is involved during the actual deployment of the robot system; however, we also put human control in the feedback loop and use them to improve the learning itself.

\textbf{\emph{{Learning from Offline Data.}}} An alternative to the human-in-the-loop paradigm is to learn from fixed robot datasets via imitation learning~\cite{pomerleau1989alvinn,zhang2017deep,mandlekar2020gti,florence2021implicit} or offline reinforcement learning (offline RL)~\cite{levine2020offline,fujimoto2019bcq,Kumar2020ConservativeQF,kidambi2020morel,yu2020mopo,yu2021combo,mandlekar2020iris,Kostrikov2021OfflineRL}.
Offline RL algorithms, particularly, have demonstrated promise when trained on large diverse datasets with suboptimal behaviors~\cite{singh2020cog,kumar2022when,ajay2020opal}.
Among a number of different methods, advantage-weighed regression methods~\cite{wang2020critic,nair2021awac,Kostrikov2021OfflineRL} have recently emerged as a popular approach to offline RL.
These methods use a weighted behavior cloning objective to learn the policy, using learned advantage estimates as the weight.
In this work, we also use weighted behavior cloning; however, we explicitly leverage human intervention signals from our online human-in-the-loop setting to obtain weights rather than using task rewards to learn advantage-based weights.
We show that this leads to superior empirical performance for our manipulation tasks.

\section{Background and Overview}

\subsection{Problem Formulation}

We formulate a robot manipulation task as a Markov Decision Process $\mathcal{M} = (\mathcal{S}, \mathcal{A}, \mathcal{R}, \mathcal{P}, p_{0}, \gamma )$ representing the state space, action space, reward function, transition probability, initial state distribution, and discount factor.
In this work, we adopt an intervention-based learning framework in which the human can choose to intervene and take control of the robot.
Given the current state $s_t\in\mathcal{S}$, the robot action $a_t^{R}\in\mathcal{A}$ is drawn from the policy $\pi_{R}\left(\cdot \mid s_{t}\right)$, and the human can override this action with a human action $a^H_t\in\mathcal{A}$.
The policy $\pi$ for the human-robot team can thus be formulated as:
\begin{equation*}
    \pi(\cdot\mid s_{t}) = I_H(s_t) \pi_H(\cdot\mid s_{t}) + (1 - I_H(s_t))\pi_R(\cdot\mid s_{t}),
\end{equation*}
where $I_H$ is a binary indicator function of human interventions and $\pi_H$ is the implicit human policy.
Our learning objective is two-fold: 1) we want to improve the level of robot autonomy by finding the autonomous policy $\pi_{R}$ that maximizes the cumulative rewards $\mathbb{E}_{\pi_{R}}\left[\sum_{t=0}^{\infty} \gamma^{t}  r\left(s_{t}, a_{t}, s_{t+1}\right)\right]$, and 2) we want to minimize the human's workload in the system, \textit{i.e.}, the expectation of interventions $\mathbb{E}_{\pi}[I_H(s_t)]$ under the state distribution induced by the team policy $\pi$.

\begin{figure}[t]
    \centering
    \includegraphics[width=1.0\linewidth]{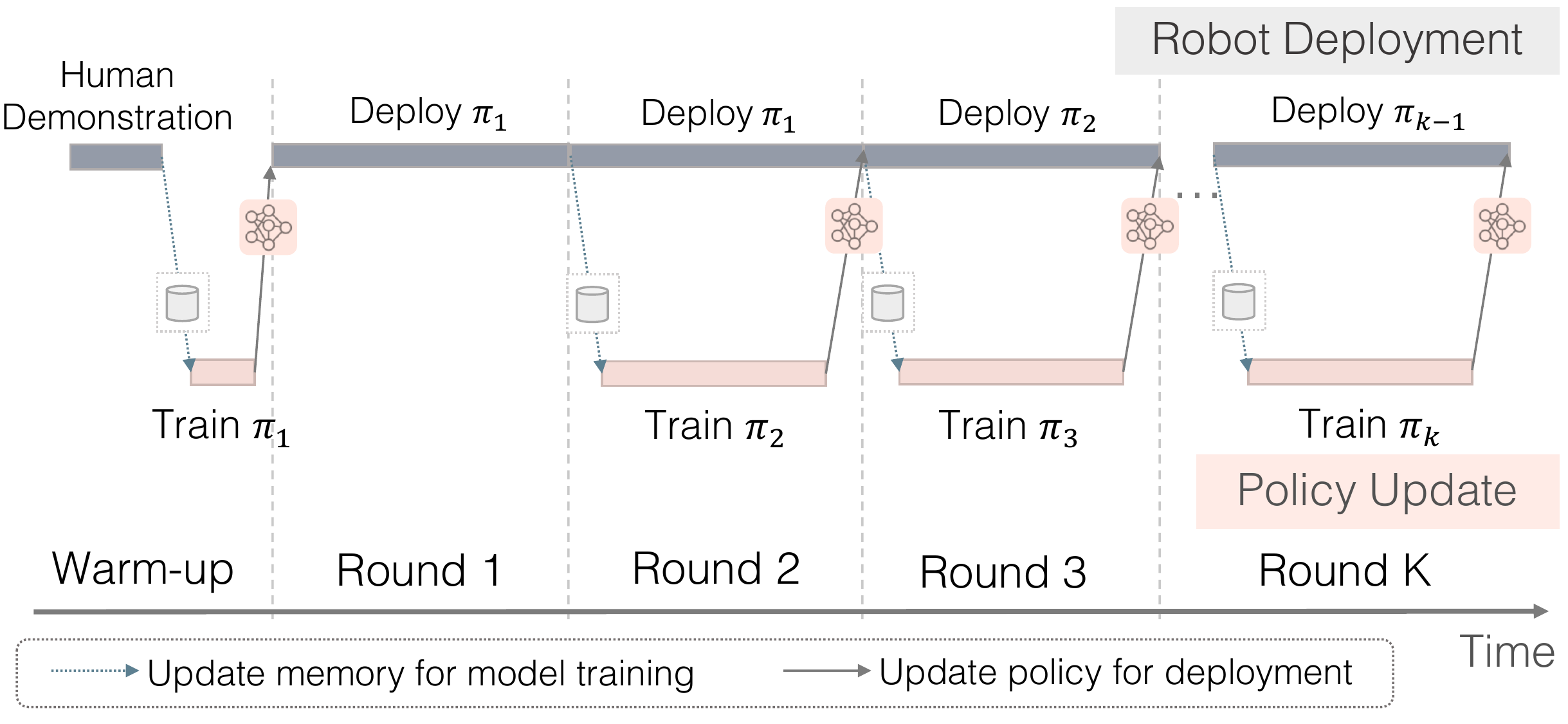}
    \caption{\textbf{Illustration of the workflow in Sirius.} Robot deployment and policy update co-occur in two parallel threads. Deployment data are passed to policy training, while a newly trained policy is deployed to the target environment for task execution.}
    \label{fig:cicd}
\end{figure}

\subsection{Weighted Behavioral Cloning Methods}

We aim to learn a robot policy $\pi_R$ with the deployment data to enhance robot autonomy and reduce human costs in human-robot collaboration. 
Weighted Behavioral Cloning (BC) has recently become one promising approach to learning policies from multimodal and suboptimal data. 
In standard BC methods, we train a model to mimic the action for each state in the dataset. The objective is to learn a policy $\pi_{R}$ parameterized by  $\theta$ that maximizes the log-likelihood of actions $a$ conditioned on the states $s$:

\begin{equation}
\theta^*=\underset{\theta}{\arg \max}  \underset{(s, a) \sim \mathcal{D}}{\mathbb{E}}\left[\log \pi_{\theta}(a \mid s)\right],
\end{equation}
where $(s,a)$ are samples from the dataset $\mathcal{D}$. For weighted BC, the log-likelihood term of each $(s, a)$ pair is scaled by a weight function $w(s,a)$, which assigns different importance scores to different samples:
\begin{equation}
\theta^*=\underset{\theta}{\arg \max}  \underset{(s, a) \sim \mathcal{D}}{\mathbb{E}}\left[ w(s,a) \log \pi_{\theta}(a \mid s)\right].
\end{equation}

The weighted BC framework lays the foundation of several state-of-the-art methods for offline reinforcement learning (RL)~\cite{nair2021awac, Kostrikov2021OfflineRL,wang2020critic}. Different weight assignments differentiate high-quality samples from low-quality ones, such that the algorithm prioritizes high-quality samples for learning. 
In particular, advantage-based offline RL algorithms calculate weights as $w(s,a) = f(Q^\pi(s,a))$, where $f(\cdot)$ is a non-negative scalar function related to the learned advantage estimates $A^\pi(s,a)$. High-advantage samples indicate that their actions likely contribute to higher future returns and, therefore, should be weighted more. Through the sample-weighting scheme, these methods filter out low-advantage samples and focus on learning from the higher-quality ones in the dataset. 
Nonetheless, effectively learning value estimates can be challenging in practice, especially when the dataset does not cover a sufficiently wide distribution of states and actions---a challenge highlighted by prior work~\cite{gulcehre2020rl, fu2020d4rl}. In the deployment setting, the data only constitute successful trajectories  that complete the task eventually. 
Empirically, we find in Section~\ref{experiments} that the nature of our deployment data makes today's offline RL methods struggle to learn values.

In contrast to the value learning framework, some prior works \cite{mandlekar2020humanintheloop, g2022eliciting, 2022correct_me} have developed weighted BC approaches that are specialized for the human-in-the-loop setting. In particular, Mandlekar et al.~\cite{mandlekar2020humanintheloop} proposes Intervention-weighted Regression (IWR) which designs weights based on whether a sample is a human intervention. 
Inspired by these prior works, we introduce a simple yet practical weighting scheme that harnesses the unique properties of deployment data to learn performant agents.
We elaborate on our weighting scheme in the following section.

\section{Sirius: Human-in-the-loop Learning and Deployment}

We present Sirius, our human-in-the-loop framework that learns and deploys continually improving policies from human and robot deployment data.
First, we define the human-in-the-loop deployment setting and give an overview of our system design.
Next, we describe our weighting scheme, which can learn effective policies from mixed, multi-modal data throughout deployment.
Finally, we introduce memory management strategies that reduce the computational complexities of policy learning and improve the efficiency of the system.

\subsection{Human-in-the-loop Deployment Framework}


Our human-in-the-loop system aims to constantly learn from the deployment experience and human corrective feedback so as to obtain a high-performing robot policy and reduce human workload over time. It consists of two components that happen simultaneously: Robot Deployment and Policy Update. In Robot Deployment (top thread in Fig. \ref{fig:cicd}), the robot performs task executions with human monitoring; in Policy Update (bottom thread), the system improves the policy with the deployment data for the next round of task execution.

The system starts with an initial policy in the warm-up phase, where we bootstrap a robot policy $\pi_{1}$ trained on a small number of human demonstrations. 
Initially, the memory buffer comprises a set of human demonstration trajectories $\mathcal{D}^0 = \{\tau_j\}$, where each trajectory $\tau_j = \{s_t, a_t, r_t, c_t=\texttt{demo}\}$ consists of the states, actions, task rewards, and the data class type flag $c_t$ indicating that these trajectories are human demonstrations. 

Upon training the initial policy $\pi_{1}$, we deploy the robot to perform the task, and in the process, we collect a set of trajectories to improve the policy.
A human operator who continuously monitors the robot's execution will intervene based on whether the robot has performed or will perform suboptimal behaviors.
Note that we adapt human-gated control~\cite{kelly2019hg} rather than robot-gated control~\cite{Hoque2021ThriftyDAggerBN}  to guarantee task execution success and trustworthiness of the system for real-world deployment.
Through this process, we obtain a new dataset $\mathcal{D}^\prime$ of trajectories $\tau_j = \{s_t, a_t, r_t, c_t\}$, where $c_t$ either indicates the transition is a \textit{robot action} ($c_t = \texttt{robot}$) or a \textit{human intervention} ($c_t = \texttt{intv}$). We append this data to the existing memory buffer collected so far $\mathcal{D}^1 \leftarrow \mathcal{D}^{0} \cup \mathcal{D}^\prime$, and train a new policy $\pi_{2}$ on this new dataset. 

In subsequent rounds, we deploy the robot to collect new data while simultaneously updating the policy. We define ``Round'' as the interval for policy update and deployment: It consists of the completion of training for one policy, and at the same time, the collection of one set of deployment data. In Round $i$, we train for policy $\pi_{i}$ using all previous data. Meanwhile, the robot is continuously being deployed using the current best policy $\pi_{i-1}$, and gathered deployment data $\mathcal{D}^\prime$.
At the end of round $i$ we append this data to the existing memory buffer collected so far $\mathcal{D}^i \leftarrow \mathcal{D}^{i-1} \cup \mathcal{D}^\prime$ and train a new policy $\pi_{i+1}$ on this aggregated dataset.

Our system aggregates data from deployment environments over long-term deployments. 
This presents a unique set of challenges: first, the generated data comes from \textbf{mixed distributions} consisting of robot policy actions, human interventions, and human demonstrations; also, the system  produces data that is \textbf{constantly growing in size}, imposing memory burden and computational inefficiency for learning algorithms. We address these challenges in the following sections.


\subsection{Human-in-the-loop Policy Learning}
\label{sec:intv-weight-scheme}

\begin{figure}[t]
    \centering
    \includegraphics[width=1.0\linewidth]{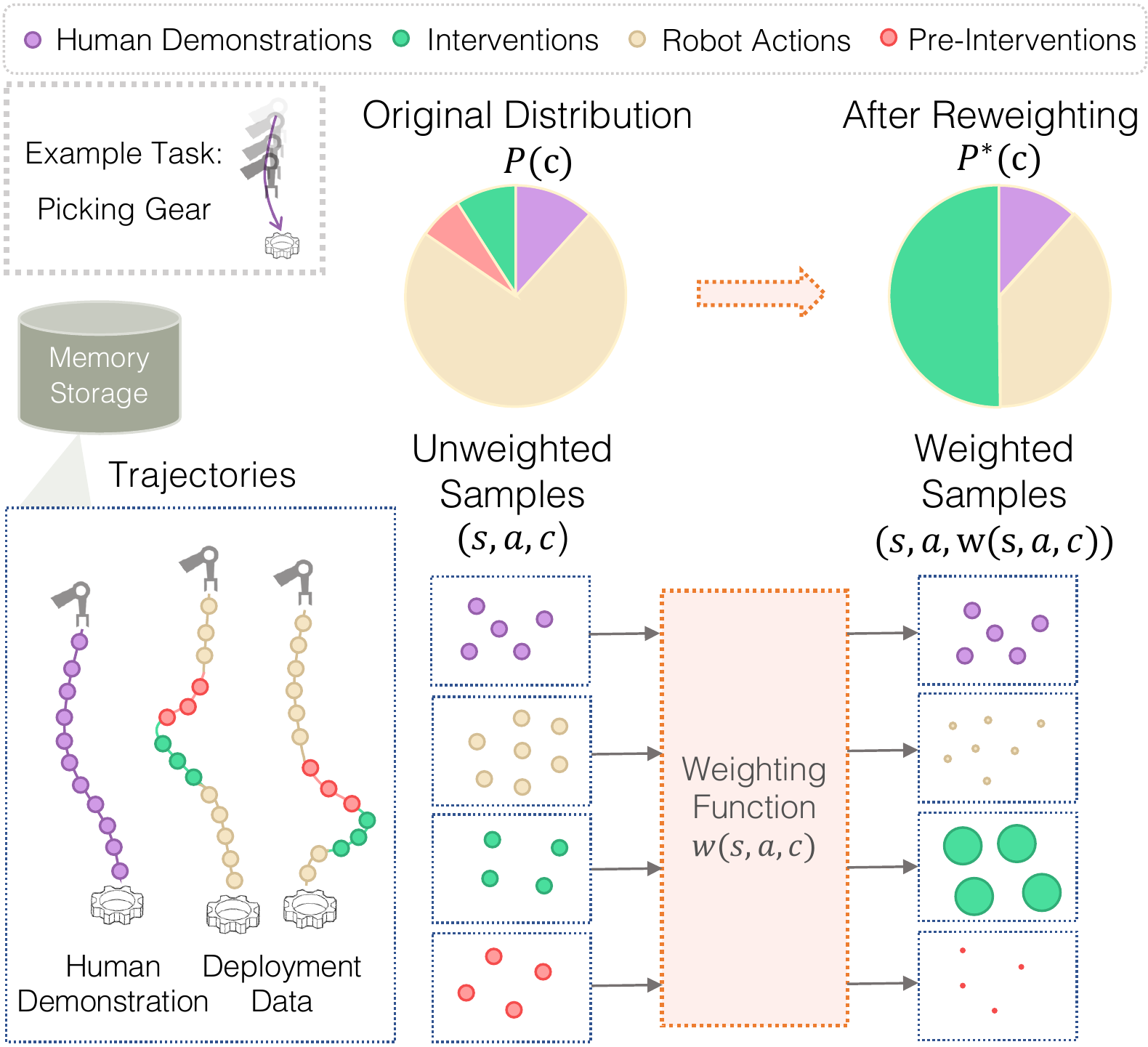}
    \caption{\textbf{Overview of our human-in-the-loop learning model.}
    We maintain an ever-growing database of diverse experiences spanning four categories: human demonstrations, autonomous robot data, human interventions, and transitions preceding interventions which we call pre-interventions. We set weights according to these four categories, with a high weight given to interventions over other categories. We use these weighted samples to continually learn vision-based manipulation policies during deployment.}
    \label{fig:model}
\end{figure}

We present a simple yet effective learning method that takes advantage of the unique characteristics of deployment data to learn effective policies.
We have a critical insight that human interventions provide informative signals of human trust and human judgement of the robot executions, which we will use to guide the design of our algorithm.
The core idea of our approach is to harness the structure
of the human correction feedback to re-weigh training samples based on an approximate quality score. With these weighted samples, we train the policy with the weighted behavioral cloning method to learn the policy on mixed-quality data. Our approach is motivated by two insights on how the human intervention structure could be used.
%


\begin{algorithm*}
\caption{Human-in-the-loop Learning at Deployment}\label{alg:algorithm}
\begin{multicols}{2}
\begin{algorithmic}
\State \textbf{Notations}
\State \hspace*{2.3mm} $L$: memory buffer maximum fixed size
\State \hspace*{2mm} $X$: maximum deployment rounds
\State \hspace*{1.7mm} $M$: number of initial human demonstration trajectories
\State \hspace*{2mm} $K$: number of rollout episodes in each deployment round
\State \hspace*{2mm} $b$: batch size
\State \hspace*{2mm} $n$: number of gradient steps in each learning round
\State \hspace*{2mm} $\alpha$: policy learning rate
\State 
\State \code{warmstart phase}
\State Collect $M$ human demonstrations $\tau_{1}, \ldots, \tau_{M}$ 
\State $\mathcal{D}^0 \leftarrow \{\tau_{1}, \ldots, \tau_{M}\} $
\State Initialize BC policy $\pi_1^{\theta}$: 
\State \hspace*{2mm} $\theta^*=\arg \max _{\theta} \mathbb{E}_{(s, a) \sim \mathcal{D}^0}\left[\log \pi^{\theta}_{1}(a \mid s)\right]$
\vspace*{0.7mm}
\State

\State \code{initial deployment data}
\State $\mathcal{D}^1 \leftarrow$ \Call{Deployment}{$\pi^{\theta}_{1}$, $\mathcal{D}^0$} 

\State
\State \code{deployment-learning loop}
\For{$i \leftarrow 1$ to $X$}
    \State Run in parallel:
    \State \hspace{2mm} $\mathcal{D}^{i+1} \leftarrow $\Call{Deployment}{$\pi^\theta_i$, $\mathcal{D}^i$}
    \State \hspace{2mm} $\pi^\theta_{i+1} \leftarrow $\Call{Learning}{$\mathcal{D}^{i}$}
\EndFor
\end{algorithmic}
\columnbreak
\begin{algorithmic}

\State \code{deployment thread}
\Function{Deployment}{$\pi_\theta$, $\mathcal{D}$} \label{f:deploy}

\State Collect rollout episodes $\tau_{1}, \ldots, \tau_{K} \sim p_{\pi_{\theta}}(\tau)$ 
\State $\mathcal{D}^{+} \leftarrow \mathcal{D} \cup\left\{\tau_{1}, \ldots, \tau_{K}\right\} $
\If{$|\mathcal{D}^{+}| > L$}
\State Discard trajectories in $\mathcal{D}^{+}$ s.t. $|\mathcal{D}^{+}| \le L$
\State \quad with a memory management strategy (in \ref{sec:memory-management}) 
\EndIf
\State
\Return $\mathcal{D}^{+}$
\EndFunction

\State
\State \code{learning thread}

\Function{Learning}{$\mathcal{D}$} \label{f:learn}
\State Initialize $\pi_\theta$
\For{each class $c$}
\State $\mathcal{D}_c \leftarrow \{(s,a,c') \in \mathcal{D} \mid c' = c\}$
\State $P(c) \leftarrow |\mathcal{D}_c| / |\mathcal{D}|$ 
\State Obtain $P^*(c)$ (see \ref{imple_details})
\EndFor

\For{$n$ gradient steps}
\State Sample mini-batch $\left(s^i, a^i, c^i\right)^b_{i=1} \sim \mathcal{D}$
\State Compute $w(s^i,a^i,c^i) \leftarrow \frac{P^*(c^i)}{P(c^i)}$ for the mini-batch
\State $\mathcal{L}_\pi(\theta) = -\frac{1}{b} \sum_i \left[w(s^i,a^i,c^i) \cdot \log \pi_{\theta}(a^i \mid s^i)\right]$
\State $\theta \leftarrow \theta \ - \ \alpha \nabla_\theta \mathcal{L}_\pi(\theta)$
\EndFor
\Return $\pi_{\theta}$
\EndFunction
\end{algorithmic}
\end{multicols}

\end{algorithm*}




Our first intuition is that human intervention samples are highly important samples and should be prioritized in learning. 
Human-operated samples are expensive to obtain and should be optimized in general, but human intervention occurs in situations
where the robot is unable to complete the task and requires help. These are risk-sensitive task states, so data in these regions are highly valuable. Therefore, these state-action pairs should be ranked high by the weighting function, and we should upweight the human intervention samples such that these samples will positively influence learning more.

Moreover, we should not only make use of \emph{what} human samples to use, but also \emph{when} the human samples take place. 
We make the critical observation that when the robot operates autonomously, it usually performs reasonable behaviors. But when it demands interventions, it is when the robot has made mistakes or has performed suboptimal behaviors.
Therefore, human interventions implicitly signify human value judgment of the robot behavior---the samples before human interventions are less desirable and of lower quality. We aim to minimize their impact on learning.

With these insights, we devise a weighting scheme according to intervention-guided data class types. Recall that each sample $(s, a, r, c)$ in our dataset contains a data class type $c$, indicating whether the sample denotes a human demonstration action, robot action, or human intervention action. To incorporate the timing of human interventions, we distinguish and penalize the samples taken prior to each human intervention. We define the segment preceding each human intervention as a separate class, pre-intervention (\texttt{preintv}) (see Fig.  \ref{fig:model}). This classification is based on the implicit human evaluation from the human partner, thresholding the robot samples into either normal \texttt{robot} samples or suboptimal \texttt{preintv} samples.
Overall, this yields four class types $c \in$ \{$\texttt{demo}$, $\texttt{intv}$, $\texttt{robot}$, $\texttt{preintv}\}$.

We derive the weight for each individual sample according to its corresponding class type $c$.
Suppose the dataset $\mathcal{D}$ has total number of samples $N$, and $n_c$ is the number of samples that is class $c$. We use $\mathcal{D}_c$ to represent the collection of samples of class $c$ in $\mathcal{D}$. The original class distribution is $P(c) = n_c / N$ for class $c$, and the unweighted BC objective under this distribution is:
\begin{equation} \label{eq:original-dist}
\begin{split}
& \hspace{5mm} \underset{\theta}{\arg \max} \  \underset{(s, a) \sim \mathcal{D}}{\mathbb{E}}\left[\log \pi_{\theta}(a \mid s)\right] \\ 
& = \underset{\theta}{\arg \max} \ \underset{P(c)}{\mathbb{E}} \  \underset{(s, a) \sim \mathcal{D}_c}{\mathbb{E}} \left[\log \pi_{\theta}(a \mid s)\right]. 
\end{split}
\end{equation}
In a long-term deployment setting, most data will be robot actions, and human interventions usually constitute a small ratio of the dataset samples, since interventions only happen at critical regions in a trajectory; the pre-intervention samples constitute a small but non-negligible proportion which can have detrimental effects (see Fig. \ref{fig:model}, left pie chart). We will now change the class distribution to a new distribution $P^*(c)$, in which we increase the ratio of human intervention samples and decrease the ratio of the pre-intervention samples (see Fig. \ref{fig:model}, right pie chart).
Under this new distribution, the weight $w(s,a,c)$ of the training samples in each individual class $c$ can be equivalently set as $w(s,a,c) = P^*(c) / P(c)$ by the rule of importance sampling. 
We outline the details of our specific distribution $P^*(c)$ in Sec. \ref{imple_details}.
This way, we obtain the sample weights for weighted BC, leveraging the inherent structure of human-robot team data.

\subsection{Memory Management}
\label{sec:memory-management}
As the deployment continues and the dataset increases, large data slows down training convergence and takes up excessive memory space. We hypothesize that forgetting (routinely discarding samples from memory) helps prioritize important and useful experiences for learning, speeding up convergence and even further improving policy. Moreover, the right kind of forgetting matters, since we want to preserve the data that is most beneficial to learning. Therefore, we would like to investigate the following question---with limited data storage and a never-ending deployment data flow, how do we absorb the most useful data and preserve more valuable information for learning? 

We assume that we have a fixed-size memory buffer that replaces existing samples with new ones when full.
We consider five strategies for managing the memory buffer of deployment data. Each strategy tests out a different hypothesis listed below: 
\begin{enumerate}
    \item \textbf{LFI} (Least-Frequently-Intervened): first reject samples from trajectories with the least interventions. \\
    \emph{(Preserving the most human intervened trajectories keeps the most valuable human and critical state examples, which helps learning the most.)}
    \item \textbf{MFI} (Most-Frequently-Intervened): first reject samples from trajectories with the most interventions. \\
    \emph{(Successful, unintervened robot trajectories yield higher quality data for learning compared to those that require intervention.)}
    \item \textbf{FIFO} (First-In-First-Out): reject samples in the order that they were added to the buffer. \\
    \emph{(More recent data from a higher performing policy
    are higher quality data for learning.)}
    \item \textbf{FILO} (First-In-Last-Out): reject the most recently added samples first. \\
    \emph{(Initial data from a worse performing policy have greater state coverage and data diversity for learning.)}
    \item \textbf{Uniform}: reject samples uniformly at random. \\
    \emph{(Uniformly selecting trajectories can yield a balanced mix of diverse samples, aiding in the learning process.)}
    
\end{enumerate}

With the intervention-guided weighting scheme for policy update and memory management strategies, we present the overall workflow of human-in-the-loop learning in deployment in Algorithm \ref{alg:algorithm}.


\begin{figure}[t]
    \centering
    \includegraphics[width=1.0\linewidth]{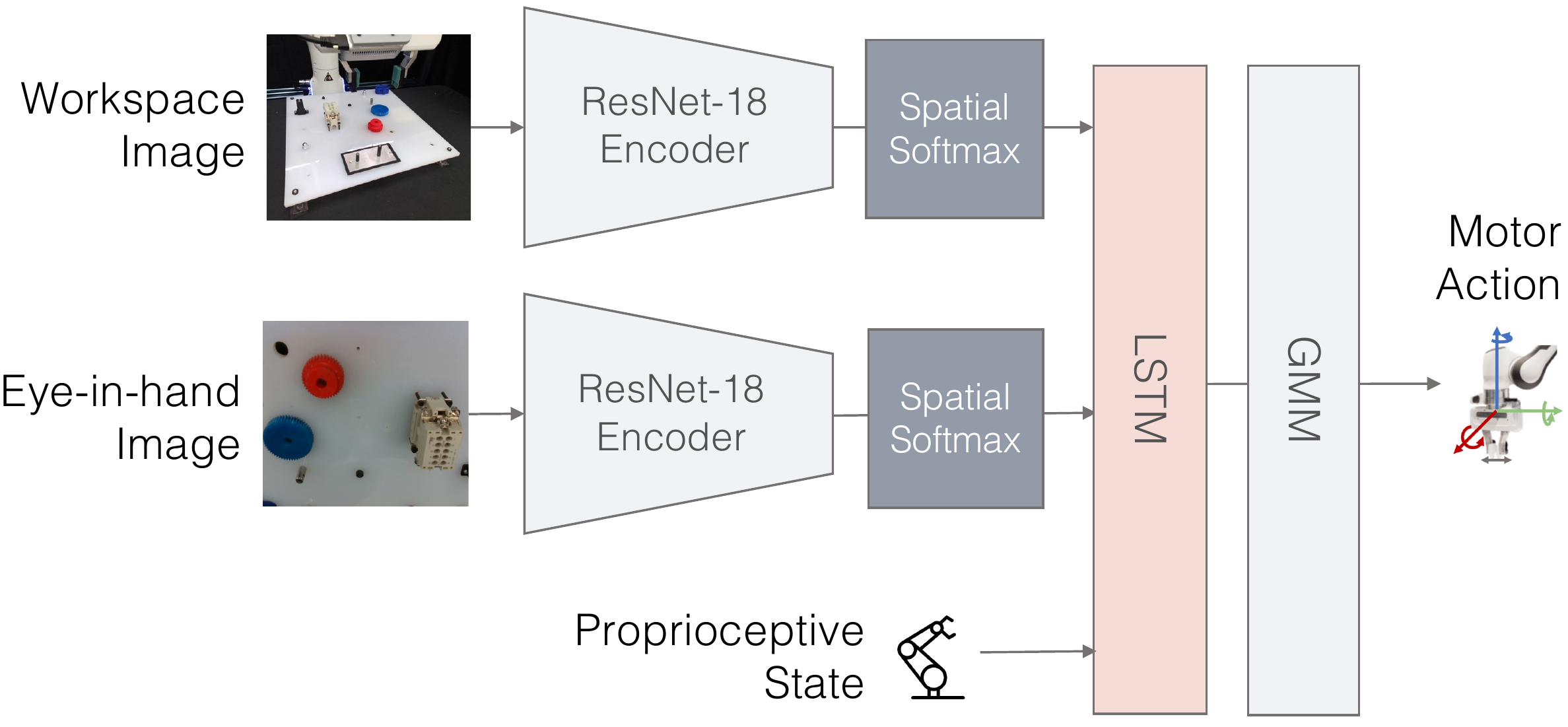}
    \caption{\textbf{Policy Architecture.} Our vision-based policy uses BC-RNN as our policy backbone. Our inputs are workspace camera image and eye-in-hand camera image, as well as robot proprioceptive states.}
    \label{fig:policy-archi}
\end{figure}

\subsection{Implementation Details} \label{imple_details}
For the robot policy (see Fig. \ref{fig:policy-archi}), we adopt BC-RNN~\cite{robomimic2021}, the state-of-the-art behavioral cloning algorithm, as our model backbone.
We use ResNet-18 encoders \cite{DBLP:journals/corr/HeZRS15} to encode third person and eye-in-hand images \cite{robomimic2021, mandlekar2020gti}.
We concatenate image features with robot proprioceptive state as input to the policy.
The network outputs a Gaussian Mixture Model (GMM) distribution over actions.

For our intervention-guided weighting scheme, we set $P^*(\texttt{intv}) = \frac{1}{2}$. The $50\%$ ratio is adapted from prior work \cite{mandlekar2020humanintheloop} that increases the weight of intervention to a reasonable level. We conduct an ablation study in Section \ref{experiments} how changing $P^*(\texttt{intv})$ affects the policy performance. We set $P^*(\texttt{preintv}) = 0$, essentially nullifying the impact of pre-intervention samples. 
The \texttt{demo} weight maintains the true ratio of demonstration samples in the dataset: $P^*(\texttt{demo}) = P(\texttt{demo})$. Finally, $P^*(\texttt{robot})$ adjusts itself accordingly.
Under this new distribution, we implicitly decrease the proportion of the \texttt{robot} class (see Fig. \ref{fig:model}) due to increasing the proportion of the \texttt{intv} class.
Note that the ratio of the demonstration remains unchanged as they are still important and useful samples to learn from, especially during initial rounds of updates when the robot generates lower-quality data.
This is in contrast to IWR by Mandlekar et al.~\cite{mandlekar2020humanintheloop}, which treats all non-intervention samples as a single class, thus lowering the contribution of demonstrations from their unweighted ratio. The weight for each individual sample is $w(s,a,c) = P^*(c) / P(c)$, as discussed in Section \ref{sec:intv-weight-scheme}.



We set a segment of length $\ell$ before each human intervention as the class \texttt{preintv}. The optimal choice on the hyperparameter $\ell$ depends on the \emph{human reaction time}, which quantifies how fast the human operator reacted to the robot’s undesired behavior. Prior works \cite{hri_response_2022, spencer2020wil} indicate that a response delay exists between the time the robot starts to perform mistakes and the time human actually perform corrective interventions. Our empirical observation based on our human operator shows an average reaction time of $2$ seconds, roughly corresponding to the time of $15$ robot actions. We thus set $\ell = 15$.


\section{Experiments}
\label{experiments}

In our experiments, we seek to answer the following research questions: 1) How effective is Sirius in improving autonomous robot policy performance over time? 2) Can this system reduce human workload over time? 3) How do the individual design choices in our learning algorithm affect overall performance? and 4) Which memory management strategy is most effective for learning with constrained memory storage?

\begin{figure*}[t]
    \centering
    \includegraphics[width=1.0\linewidth]{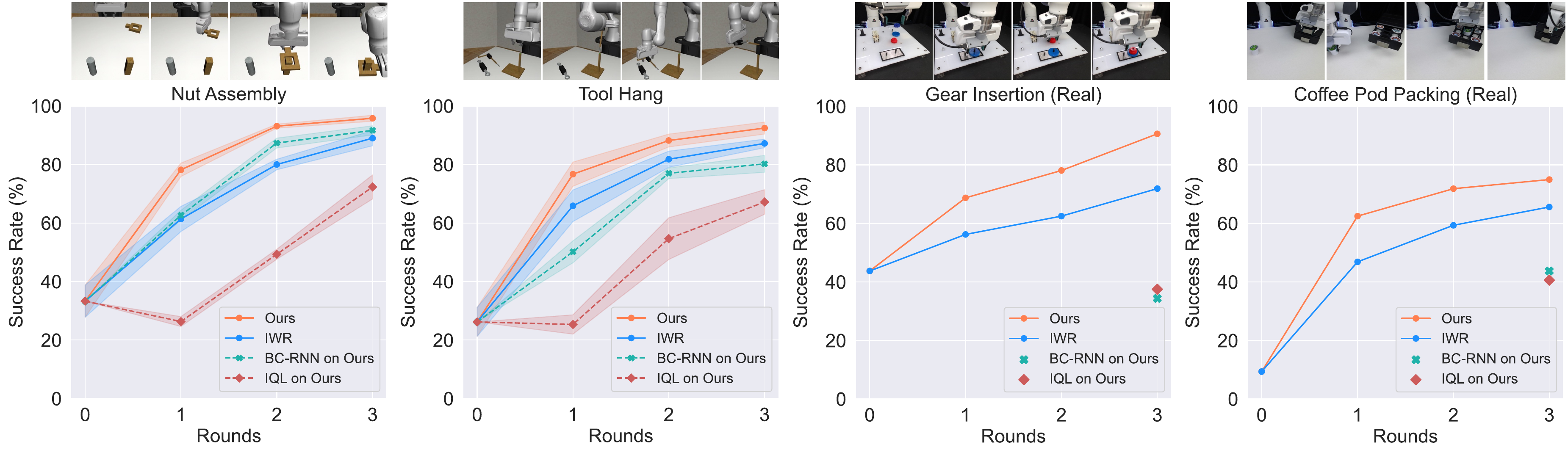}
    \caption{\textbf{Quantitative evaluations.} We compare our method with human-in-the-loop learning, imitation learning, and offline reinforcement learning baselines. Our results in simulated and real-world tasks show steady performance improvements of the autonomous policies over rounds. Our model reports the highest performance in all four tasks after three rounds of deployments and policy updates. Solid line: human-in-the-loop; dashed line: offline learning on data from our method.}
    \label{fig:results}
\end{figure*}

\subsection{Tasks}

We design a set of simulated and real-world tasks that resemble common industrial tasks in manufacturing and logistics. We consider long-horizon tasks that require precise contact-rich manipulation, necessitating human guidance. For all tasks, we use a Franka Emika Panda robot arm equipped with a parallel jaw gripper. Both the agent and human control the robot in task space. We use a SpaceMouse as the human interface device to intervene.

We systematically evaluate the performance of our method and baselines in the robosuite simulator~\cite{zhu2020robosuite}.
We choose the two most challenging contact-rich manipulation tasks in the robomimic benchmark~\cite{robomimic2021}:

\vspace{1mm}
\textbf{Nut Assembly.} The robot picks up a square nut from the table and inserts the nut into a column.

\textbf{Tool Hang.} The robot picks up a hook piece and inserts it into a very small hole, then hangs a wrench on the hook. As noted in robomimic~\cite{robomimic2021}, this is a difficult task requiring precise and dexterous control.
\vspace{1mm}

In the real world, we design two tasks representative of industrial assembly and food packaging applications:

\vspace{1mm}
\textbf{Gear Insertion.} The robot picks up two gears on the NIST board and inserts each of them onto the gear shafts.

\textbf{Coffee Pod Packing.} The robot opens a drawer, places a coffee pod into the pod holder, and closes the drawer. 

\begin{figure}[t]
    \centering
    \includegraphics[width=1.0\linewidth]{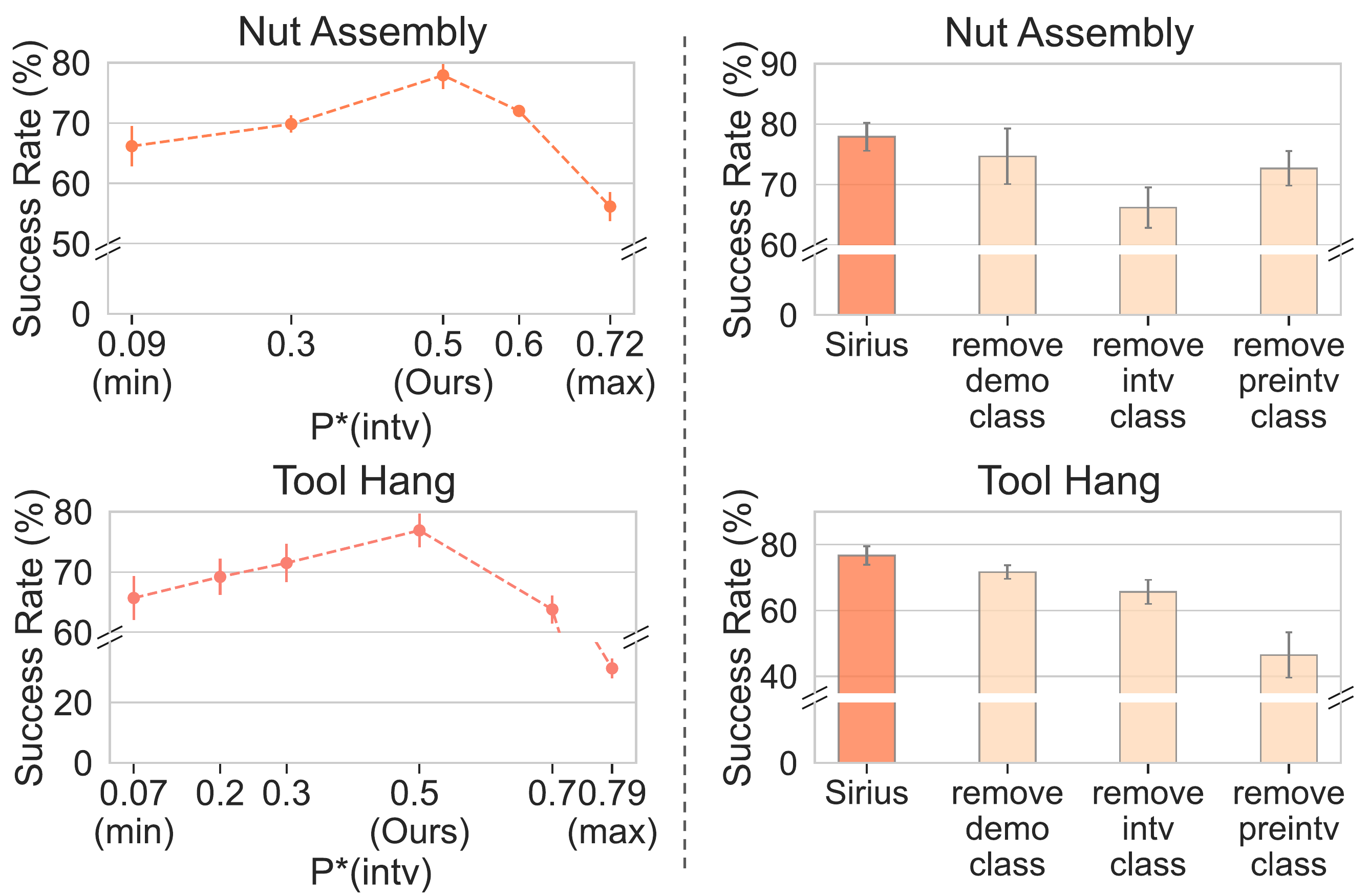}
    \caption{\textbf{(Left) Ablation on intervention ratio weight.} We show how policy performance first increase then decrease as $P^*(\texttt{intv})$ increases, pearking at $P^*(\texttt{intv}) = 0.5$. \textbf{(Right) Ablation on weight function design.} Our results show that removing each class label hurts model performance. }
    \vspace{-0.5mm}
    \label{fig:ablation_studies}
\end{figure}

\subsection{Baselines and Evaluation Protocol}
\label{metrics}

We compare our method with the state-of-the-art human-in-the-loop learning method for robot manipulation, Intervention Weighted Regression (\textbf{IWR})~\cite{mandlekar2020humanintheloop}. Furthermore, to ablate the impacts of algorithms versus data distributions, we compare the state-of-the-art imitation learning algorithm \textbf{BC-RNN}~\cite{robomimic2021} and offline RL algorithm Implicit Q-Learning (\textbf{IQL})~\cite{Kostrikov2021OfflineRL}. We run these two latter baselines on the deployment data generated by our method for a fair comparison.

To mimic the intervention-guided weights for IQL, we use the following rewards after hyperparameter optimization: $r=1.0$ upon task success, $r=0.25$ for intervention states, $r=-0.25$ for pre-intervention states, and $r=0$ for all other states. We also run IQL in a sparse reward setting but find it underperformed. Note that in contrast to our method, IQL requires additional information on task rewards, which may be expensive to obtain in real-world settings.

To provide a fair comparison with existing human-in-the-loop methods, we follow the round update protocol established by prior work~\cite{mandlekar2020humanintheloop, kelly2019hg}: three rounds of policy learning and deployment, where each round deployment runs until the number of intervention samples reaches one third of the initial human demonstration samples.

We benchmark human-in-the-loop deployment systems in two aspects: 1) \textbf{Policy Performance.} Our human-robot team achieves a reliable task success of 100\%. Here we evaluate the success rate of the autonomous policy after each round of model update; and 2) \textbf{Human Workload.} 
We measure human workload as the percentage of intervention in the trajectories in each round.
We perform rigorous evaluations of policy performance as follows:

\begin{itemize}
    \item \textit{Simulation experiments}: We evaluate the success rate of each method across 3 seeds. For each seed, we evaluate the success rate at a set of regularly spaced training checkpoints and record the average over the top three performing checkpoints to avoid outliers. For each checkpoint, we evaluate whether the agent successfully completed the task over 100 trials.
    \item \textit{Real-world experiments}: We evaluate each method for one seed due to the high time cost for real robot evaluation. Since real robot evaluations are subject to noise and variation across checkpoints, we first perform an initial evaluation of different checkpoints (5 checkpoints) for each method, evaluating each of them for a small number of trials (5 trials). For the checkpoint that gives the best initial quantitative behavior, we perform 32 trials and report the success rate over them.
\end{itemize}

\subsection{Experiment Results}
\label{quant_results}

\textbf{Quantitative Results.} We show in Fig.~\ref{fig:results} that our method significantly outperforms the baselines on our evaluation tasks. Our method consistently outperforms IWR over the rounds. We attribute this difference to our fine-grained weighting scheme, enabling the method to better differentiate high-quality and suboptimal samples.
This advantage over IWR cascades across the rounds, as we obtain a better policy, which in turn yields better deployment data.

We also show that our method significantly outperforms the BC-RNN and IQL baselines under the same dataset distribution. This highlights the importance of our weighting scheme --- BC-RNN performs poorly due to copying the suboptimal behaviors in the dataset, while IQL fails to learn values as weights that yield effective policy performance.

\begin{figure}[t]
    \centering
    \includegraphics[width=1.0\linewidth]{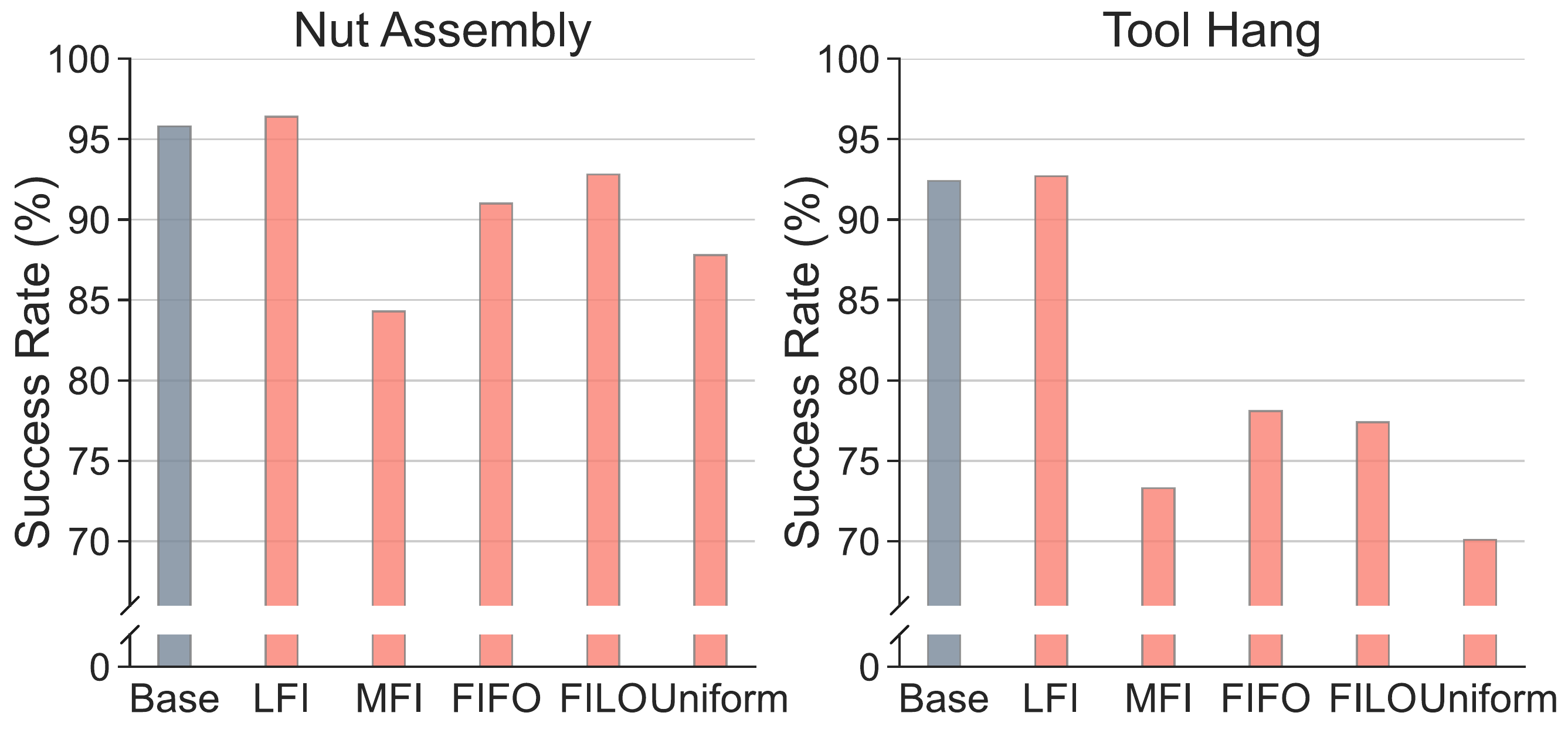}
    \caption{\textbf{Ablation on memory management strategies.} We study the five different strategies introduced in Section~\ref{sec:memory-management}. LFI (discarding least frequently intervened trajectories) matches and even yields better performance over keeping all data samples (Base) while taking much less memory storage.}
    \vspace{3mm}
    \label{fig:memory}
\end{figure}

\textbf{Ablation Studies.}
We perform an ablation study to examine the contribution of each component in our weighting scheme in Fig. \ref{fig:ablation_studies} (Right). We study how removing each class, \textit{i.e.}, treating each class as the robot action class (and thus removing the special weight for that class), affects the policy performance:
\begin{itemize}
    \item \textbf{remove demo class:} not preserving the true ratio of \texttt{demo} class, which lowers its contribution (see \ref{imple_details}).
    \item \textbf{remove intv class:} not upweighting the \texttt{intv} class, which is equivalent to (min) in Fig. \ref{fig:ablation_studies} (Left).
    \item \textbf{remove preintv class:} not downweighting the \texttt{preintv} class but treating it as \texttt{robot} class.
\end{itemize}
We run each ablated version of our method on Round $1$ data for the simulation tasks. We choose Round $1$ data for this study because they are generated from the initial BC-RNN policy rather than biased toward data generated from our method. As shown in Fig. \ref{fig:ablation_studies} (Right), removing any class weight hurts the policy performance. This shows the effectiveness of our fine-grained weighting scheme, where each class contributes differently to the learning of the deployment data.

We also conduct an in-depth study on the influence of human intervention reweighting ratio $P^*(\texttt{intv})$. In the unweighted distribution, the human intervention samples take up a small proportion of the dataset size, which we denote as the \textit{minimum ratio}; the maximum ratio it can take is to nullify the proportion of \texttt{robot} samples altogether (so that the dataset only constitutes human demonstrations and human interventions). We run our method with a different ratio ranging from minimum to maximum using Round $1$ data on both simulation tasks. The specific range for Nut Assembly and Tool Hang can be found in Fig. \ref{fig:ablation_studies} (Left). The overall trend is that the policy performance peaks at $P^*(\texttt{intv}) = 0.5$, and is worse when $P^*(\texttt{intv})$ gets larger or smaller. Our intuition is that if the intervention ratio is too small, we are not making the best use of the intervention samples; if it is too large, it will limit the diversity of training data. Either way has an adversarial effect. 

\begin{figure}[t]
    \centering
    \includegraphics[width=1.0\linewidth]{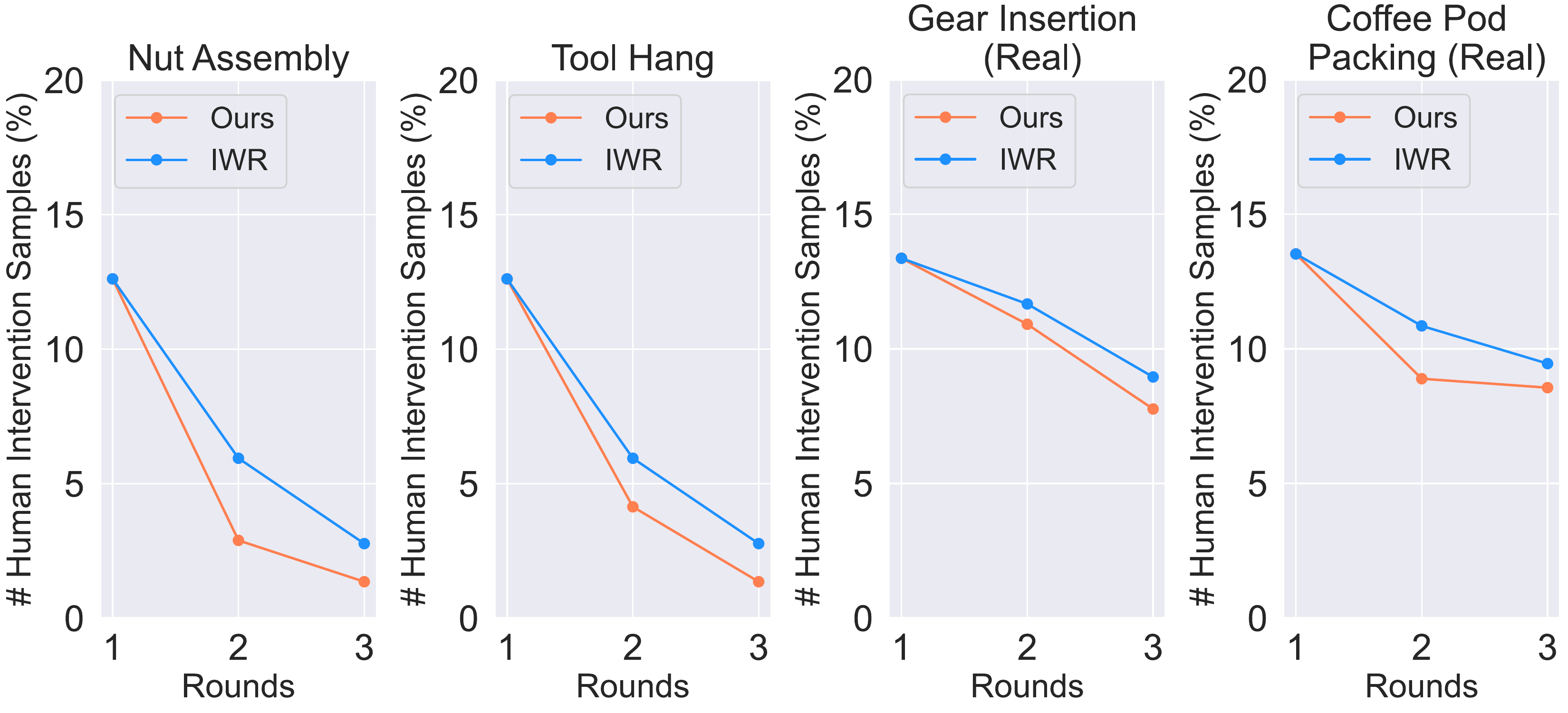}
    \caption{\textbf{Human Intervention Sample Ratio.} We evaluate the human intervention sample ratio for the four tasks. The human intervention sample ratio decreases over deployment round updates. Our methods have a larger reduction in human intervention ratio as compared with IWR.}
    \label{fig:intv_sample_ratio}
\end{figure}

\textbf{Analysis on Memory Management.} We compare the effectiveness of Memory Management strategies in Section~\ref{sec:memory-management} at deployment. Fig. \ref{fig:memory} shows the result of memory size reduction on the two simulation tasks in Round $3$, where the Nut Assembly accumulated $3000+$ trajectories and the Tool Hang task $1600+$ trajectories. By capping our memory buffer size at $500$ trajectories, we manage to reduce memory size to a much small proportion of the original dataset size (15\% for Nut Assembly and 30\% for Tool Hang). 


Among all of the strategies, LFI (discarding least frequently intervened trajectories) is the only strategy that matches and even yields better performance over keeping all data samples (Base). In addition to minimizing storage requirements, LFI also improves learning efficiency. Under LFI, the policy converged twice as fast as Base for both tasks (where we define convergence as the number of epochs to reach 90\% success rate). The faster convergence speed, in turn, yields faster model iterations in real-world deployments.

There are a number of potential explanations for the superior performance of LFI. First, note that among all of the strategies, LFI preserves the largest number of human intervention samples. This suggests that human interventions have high intrinsic value to our learning algorithm, as they help to ensure robust policy execution under suboptimal scenarios.  Another perspective is that LFI preserves the more frequently intervened trajectories, which exhibit wider state coverage and a diverse array of events. This facilitates the trained policies to operate effectively under rare and unexpected scenarios.
MFI (discarding most intervened trajectories) has the opposite effect, favoring trajectories that require less human supervision and often exhibit less diverse behaviors.
The results on FIFO and FILO suggest that managing samples according to deployment time is not the most effective strategy, as valuable training data can be collected all throughout the deployment of the system.
Finally, the na\"ive Uniform strategy is ineffective as it does not incorporate any distinguishing characteristics of samples to manage the memory.


\begin{figure}[t]
    \centering
    \includegraphics[width=1.0\linewidth]{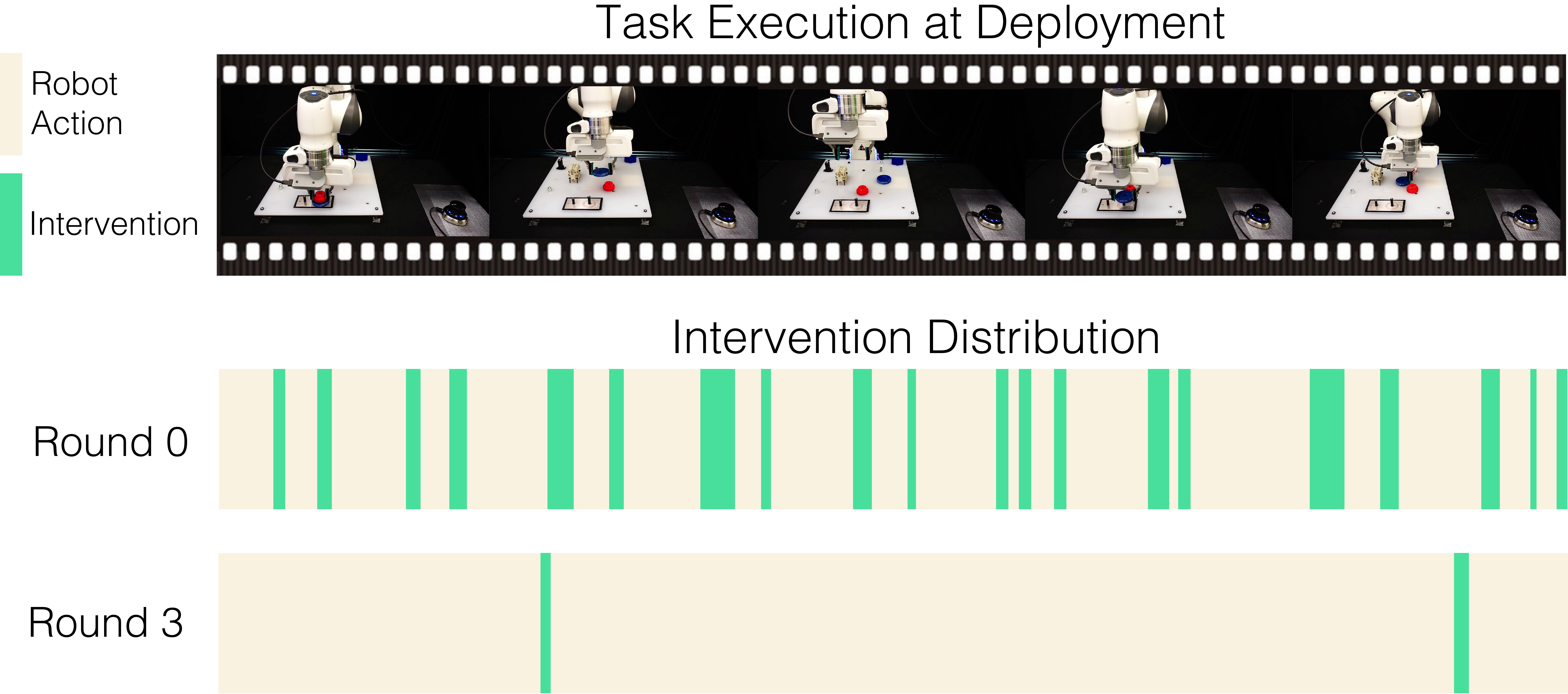}
    \caption{\textbf{Human Intervention Distribution.} The two color bars represent the time duration over 10 consecutive trajectories and whether each step is autonomous robot action (yellow) or human intervention (green). In Round 1, much human intervention is needed to handle difficult situations. In Round 3, the policy needs very little human intervention, and the robot can run autonomously most of the time.}
    \label{fig:intv-dist-gear}
\end{figure}

\textbf{Human Workload Reduction.} Lastly, we highlight the effectiveness of our method in reducing human workload. In Fig. \ref{fig:intv_sample_ratio}, we plot the human intervention sample ratio for every round, \textit{i.e.}, the percentage of intervention samples in all samples per round. We compare the results for the HITL methods, Ours and IWR. We see that the human intervention ratio decreases over rounds for both methods, as policy performance increases over time. Furthermore, we see that this reduction in human workload is greater for our method compared to IWR.

Qualitatively, we visualize how the division of work of the human-robot team evolves in Figure \ref{fig:intv-dist-gear}. For the Gear Insertion task, we do $10$ trials of task execution in sequence for our method in Round $0$ and Round $3$, respectively, and record the time duration for human intervention needed during the deployment. Comparing Round $0$ and Round $3$, the policy in Round $3$ needs very little human intervention, and the intervention duration is also much shorter. This confirms the effectiveness of our framework in human workload reduction.

\textbf{Limitations.} Our human-in-the-loop experiment of each task is only conducted with a single human operator. The results can be biased toward the individual's skills, familiarity with the system, and level of risk tolerance. A more extensive human study would enhance our understanding of how human's trust and subjectivity are manifested in time, criteria, and duration of interventions. Furthermore, to ensure trustworthy execution, our current system still requires the human to constantly monitor the robot. Incorporating automated runtime monitoring and error detection strategies~\cite{ensembledagger, Hoque2021ThriftyDAggerBN} would further reduce the human's mental burden. Lastly, for the study of human workload reduction, we employed a simplistic way of measuring human workload based on the percentage of intervention. Conducting in-depth human studies to measure human mental workload would provide deeper insights.

\vspace{1.5mm}
\section{Conclusion} 
\vspace{1mm}
\label{sec:conclusion}
We introduce Sirius, a framework for human-in-the-loop robot manipulation and learning at deployment that both guarantees reliable task execution and also improves autonomous policy performance over time.
We utilize the properties and assumptions of human-robot collaboration to develop an intervention-based weighted behavioral cloning method for effectively using deployment data. We also design a practical system that trains and deploys new models continuously under memory constraints.
For future work, we would like to improve the flexibility and adaptability of the human-robot shared autonomy, including more intuitive control interfaces and faster policy learning from human feedback. Another direction for future research is alleviating the human cognitive burdens of monitoring and teleoperating the system. Deployment monitoring would be an exciting research direction, allowing the system to automatically detect robot errors without constant human supervision.
\section*{ACKNOWLEDGMENT}
We thank Ajay Mandlekar for having multiple insightful discussions, and for sharing well-designed simulation task environments and codebases during development of the project. We thank Yifeng Zhu for valuable advice and system infrastructure development for real robot experiments. We would like to thank Tian Gao, Jake Grigsby, Zhenyu Jiang, Ajay Mandlekar, Braham Snyder, and Yifeng Zhu for providing helpful feedback for this manuscript. We acknowledge the support of the National Science Foundation (1955523, 2145283), the Office of Naval Research (N00014-22-1-2204), and Amazon.



\printbibliography

\clearpage
\section{Appendix} 
\label{sec:appendix}

\subsection{Task Details}

We elaborate on the four tasks in this section, providing more details of the task setups, the bottleneck regions, and how they are challenging. The two simulation tasks, Nut Assembly and Tool Hang, are from the robomimic codebase \cite{robomimic2021} for better benchmarking.

\textbf{Nut Assembly.} The robot picks up a square rod from the table and inserts the rod into a column. The bottleneck lies in grasping the square rod with the correct orientation and turning it such that it aims at the column correctly. 

\textbf{Tool Hang.} The robot picks up a hook piece, inserts it into a tiny hole, and then hangs a wrench on the hook. As noted in robomimic~\cite{robomimic2021}, this task requires very precise and dexterous control. There are multiple bottleneck regions: picking up the hook piece with the correct orientation, inserting the hook piece with high precision in both position and orientation, picking out the wrench, and carefully aiming the tiny hole at the hook.

\textbf{Gear Insertion.} We design the task scene setup adapting from the common NIST board benchmark\footnote{https://www.nist.gov/el/intelligent-systems-division-73500/robotic-grasping-and-manipulation-assembly/assembly} Task Board 1, which is designed for standard industrial tasks like peg insertion and electrical connector insertions. Initially, one blue gear and one red gear are placed at a randomized region on the board. The robot picks up two gears in sequence and inserts each onto the gear shafts respectively. The gears' holes are very small, requiring precise insertion on the gear shafts. 

\textbf{Coffee Pod Packing.} We design this task for a food manufacturing setting where the robot packs real coffee pods\footnote{https://www.amazon.com/gp/product/B00I5FWWPI} into a coffee pod holder\footnote{https://www.amazon.com/gp/product/B07D7M93ZW}. The robot first opens the coffee pod holder drawer, grasps a coffee pod placed on a random initial position on the table, places the coffee pod into the pod holder, and closes the drawer. The pod holder contains holes that fit precisely to the coffee pods' side, so it requires precise insertion of the coffee pods into the holes. The common bottlenecks are exactly grasping the coffee pod, exact insertion, and releasing the drawer whenever the opening and closing actions are done without getting stuck.

The objects in all tasks are initialized randomly within an x-y position range and with a rotation on the z-axis. The configurations of the simulation tasks follow that in robomimic. We present the reset initialization configuration in Table \ref{table:objects} for reference.

\subsection{Human-Robot Teaming}

\begin{figure}[t]
    \centering
    \includegraphics[width=1.0\linewidth]{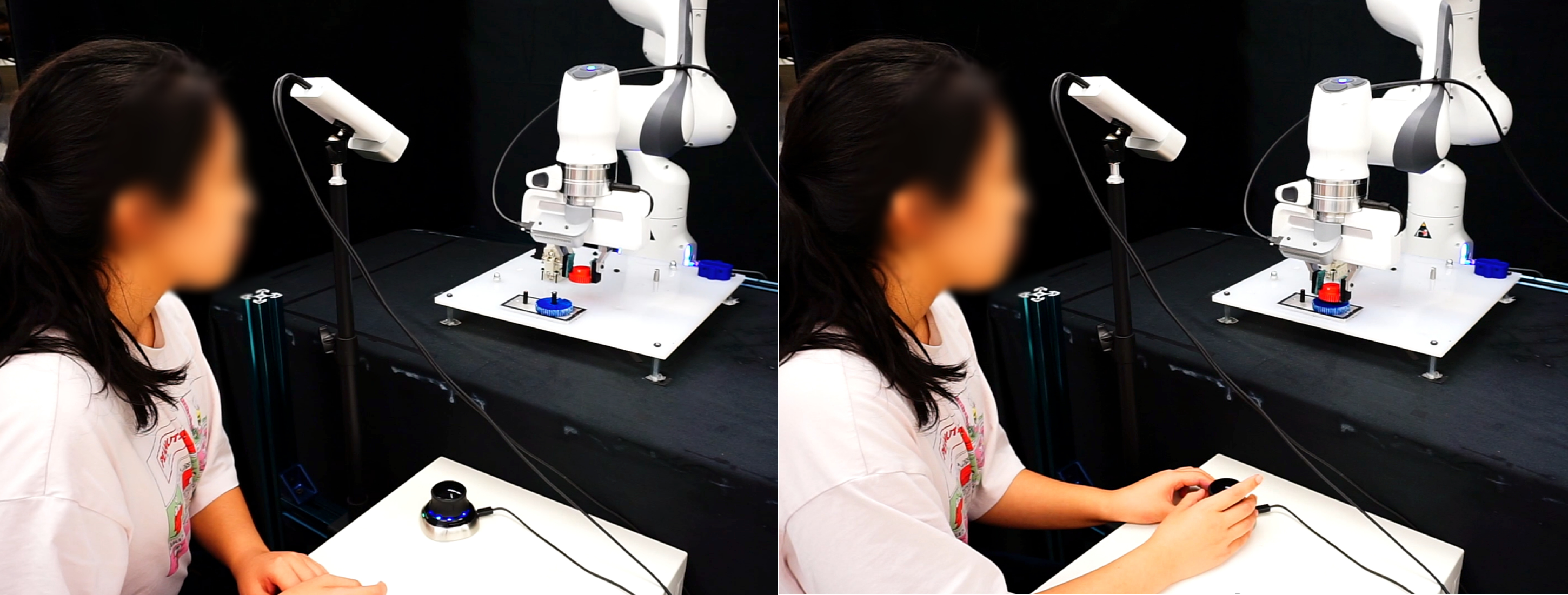}
    \caption{\textbf{Human Robot Teaming.} Left: The robot executes the task by default while a human supervises the execution. Right: When the human detects undesirable robot behavior, the human intervenes.}
    \label{fig:hri}
\end{figure}

We illustrate the actual human-robot teaming process during human-in-the-loop deployment in Figure~\ref{fig:hri}. The robot executes a task (\textit{e.g.}, gear insertion) by default while a human supervises the execution. In this gear insertion scenario, the expected robot behavior is to pick up the gear and insert it down the gear shaft. When the human detects undesirable robot behavior (\textit{e.g.}, gear getting stuck), the human intervenes by taking over control of the robot. The human directly passes in action commands to perform the desired behavior. When the human judges that the robot can continue the task, the human passes control back to the robot.

To enable effective shared human control of the robot, we seek a teleoperation interface that (1) enables humans to control the robot effectively and intuitively and (2) switches between robot and human control immediately once the human decides to intervene or pass the control back to the robot.
To this end, we employ SpaceMouse\footnote{https://3dconnexion.com/us/spacemouse/} control. The human operator controls a 6-DoF SpaceMouse and passes the position and orientation of the SpaceMouse as action commands. The user can pause when monitoring the computer screen by pressing a button, exert control until the robot is back to an acceptable state, and pass the control back to the robot by stopping the motion on the SpaceMouse.

\subsection{Observation and Action Space}

The observational space of all our tasks consists of the workspace camera image, the eye-in-hand camera image, and low-dimensional proprioceptive information. For simulation tasks, we use the operational space controller (OSC) that has a 7D action space; for real-world tasks, we use OSC yaw controller that has a 5D action space.

The minor differences for the Tool Hang task from robomimic \cite{robomimic2021} default image observation: We use an image size of $128 \times 128$ instead of the default $224 \times 224$ for training efficiency. Due to the task's need for high-resolution image inputs, we adjust the workspace camera angle to give more details on the objects. This compensates for the need for large image size and boosts policy performance.

Details on low-dimensional proprioceptive information: For simulation tasks, we have the end effector position (3D) and orientation (4D), as well as the distance of the gripper (2D). We have joint positions (7D) and gripper width (1D) for real-world tasks.

The action space of simulation tasks is 7 dimensions in total: x-y-z position (3D), yaw-pitch-roll orientation (3D), and the gripper open-close command $\{1., -1.\}$ (1D). The action space of real-world tasks is 5 dimensions in total: x-y-z position (3D), yaw orientation (1d), and the gripper open-close command $\{1., -1.\}$ (1D).

\subsection{Method Implementations}
\label{app:imple_tables}

We describe the policy architecture details initally introduced in Section \ref{imple_details}. Our codebase is based on robomimic \cite{robomimic2021}, a recent open-source project that benchmarks a range of learning algorithms on offline data. We standardize all methods with the same state-of-the-art policy architectures and hyperparameters from robomimic. The architectural design includes ResNet-18 image encoders, random cropping for image augmentation, GMM head, and the same training procedures. The list of hyperparameter choices is presented in Table \ref{table:common}. For all BC-related methods, including Ours, IWR, and BC-RNN, we use the same BC-RNN architecture specified in Table \ref{table:bc-hyper}.

For all tasks except for Tool Hang, we use the same hyperparameters with image size $84 \times 84$. We use $128 \times 128$ for Tool Hang due to its need for high-precision details. We use a few demonstrations for each task to warm-start the policy; the number ranges from $30$ to $80$ so that the initial policy can all have some level of reasonable behavior regardless of task difficulty. See Table \ref{table:task-dependent} for all task-dependent hyperparameters.

For IQL \cite{Kostrikov2021OfflineRL}, we reimplemented the method in our robomimic-based codebase to keep the policy backbone and common architecture the same across all methods. Our implementation is based on the publicly available PyTorch implementation of IQL\footnote{https://github.com/rail-berkeley/rlkit/tree/master/examples/iql}.

We follow the paper's original design with some slight modifications. In particular, the original IQL uses the sparse reward setting where the reward is based on task success. We add a denser reward for IQL to incorporate information on human intervention. To mimic the intervention-guided weights for IQL, we use the following rewards: $r=1.0$ upon task success, $r=0.25$ for intervention states, $r=-0.25$ for pre-intervention states, and $r=0$ for all other states. We found that this version of IQL outperforms the default sparse reward setting. We list the hyperparameters for IQL baseline in Table \ref{table:iql}.

\subsection{HITL System Policy Updates}
\label{app:hitl-round-update}

We elaborate on our design choice for HITL system policy update rules discussed in Section \ref{metrics} of the main paper.

In a practical human-in-the-loop deployment system, there can be many possible design choices for the condition and frequency of policy updates. A few straightforward ones among various designs are: update every specific amount of elapsed time, update after the robot completes a certain number of tasks, or update after human interventions reach a certain number. Our experiments aim to provide a fair comparison between various human-in-the-loop methods and benchmark our method against prior baselines. For consistent evaluation, we follow round updates rules by prior work~\cite{mandlekar2020humanintheloop, kelly2019hg}: 3 rounds of update when the number of intervention samples reaches $1 / 3$ of the human demonstration samples. The motivation is to evaluate prior baselines in their original setting to ensure fair comparison; moreover, we want to ensure all methods get the same amount of human samples per round. Since they are human-in-the-loop methods, the amount of human samples is important to their utilization. How policies are updated could be a dimension of human-in-the-loop system design  on its own right and could be further explored in future work.

\subsection{Human Workload Reduction}

We present more results on the effectiveness of our method in reducing human workload as discussed in the main paper. We note that there are different metrics to evaluate human workload, such as the number of control switches and lengths of interventions, as introduced in prior work~\cite{Hoque2021ThriftyDAggerBN}. We include two additional human workload metrics: 

\begin{figure}[t]
    \centering
    \includegraphics[width=1.0\linewidth]{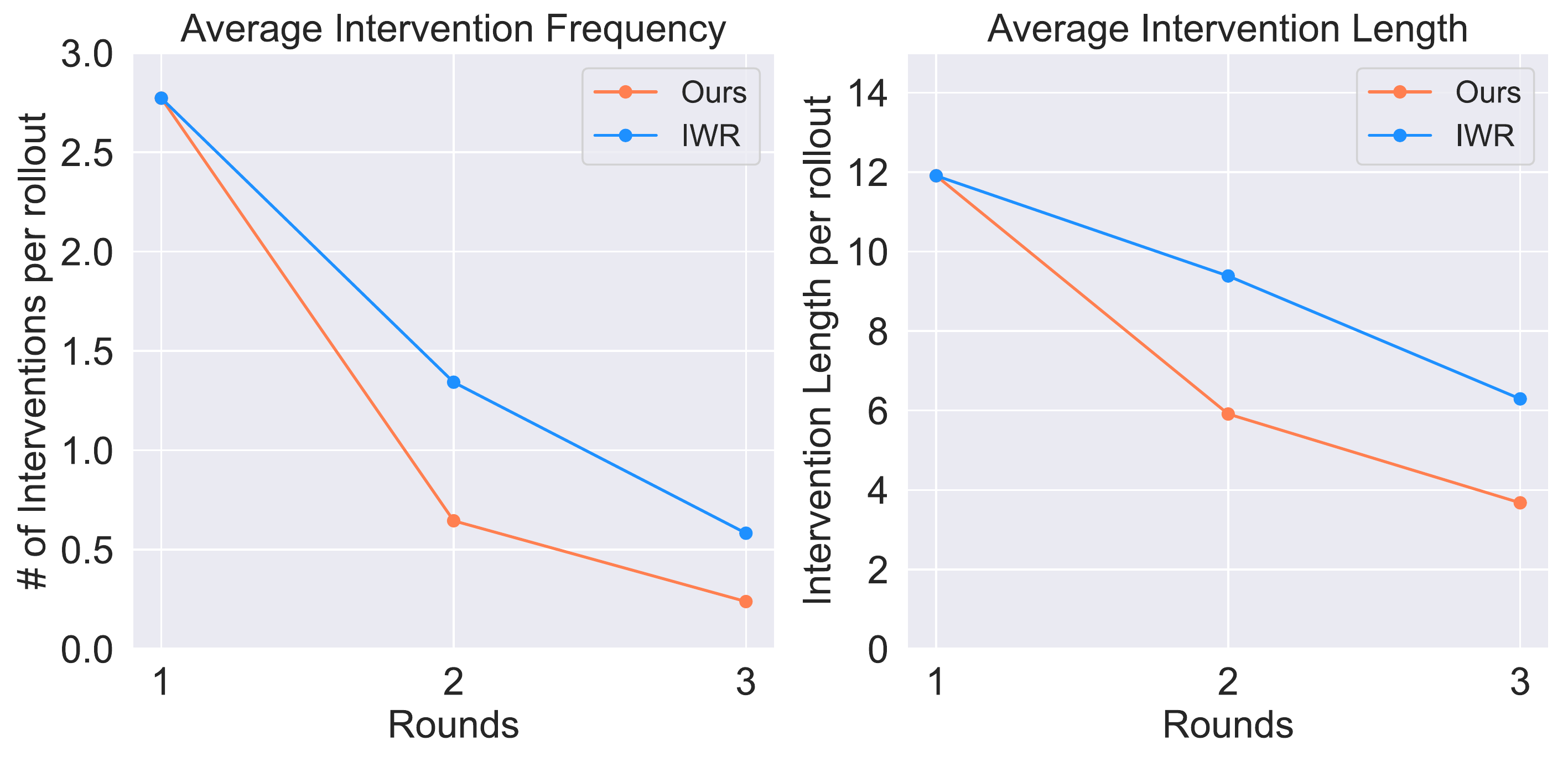}
    \caption{\textbf{More Intervention Behavior Metrics (Nut Assembly).} We present two more metrics to measure human workload over time: average intervention frequency (Left) and average intervention length (Right). We show that our method results in a larger reduction of both metrics over round updates, developing better human trust and human-robot partnership. }
    \label{fig:more-intv-metrics}
\end{figure}

\textbf{Average intervention frequency}: the number of intervention occurrences divided by the number of rollouts. This reflects the number of context switches, \textit{i.e.}, shifts of control between the human and the robot. A higher number of context switches imposes higher concentration and exhaustion on the human.

\textbf{Average intervention length}: length of each intervention in terms of the number of timesteps. This reflects the ease of every intervention - longer intervention occurrence means a higher mental workload to the human for taking control of the robot.

We note that these metrics also reflect the human trust level for the robot. The human makes a decision during robot control: should I intervene at this point? Furthermore, during human control: is the robot in a state where I can safely return control to the robot? Lower intervention frequency and shorter intervention length reflect that human trusts the robot more so that they can intervene at fewer places and return control to the robot faster.

We present the results in Figure \ref{fig:more-intv-metrics} using Nut Assembly as an example. We can see that, like the human intervention ratio, the average intervention frequency, and the intervention length decrease. Our method also has a faster reduction of both metrics over round updates. This shows that our human-in-the-loop system fosters good human trust in the robot and develops better human-robot partnerships.

\begin{table*}
\vspace{15pt}
\centering
\caption{Task objects configuration}
\begin{tabular}{c c c c c}
\hline
\textbf{Tasks and Objects} & Position (x-y) & Orientation (z) \\
\hline
& \\
\textbf{Nut Assembly} & \\
square nut & $0.5$cm $\times 11.5$cm & $2 \pi$ \\
 & \\
\textbf{ToolHang} & \\
hook & $2$cm $\times$ $2$cm & $\pi / 9$ \\
wrench & $2$cm $\times$ $2$cm & $\pi / 9$ \\
 & \\

\textbf{Gear Insertion (Real)} & \\
blue gear & $12$cm $\times$ $12$cm & $2 \pi$ \\
red gear & $12$cm $\times$ $12$cm & $2 \pi$ \\
 & \\
\textbf{Coffee Pod Packing (Real)} & \\
coffee pod & $16$cm $\times$ $16$cm & $2 \pi$ \\
& \\
\hline
\end{tabular}
\label{table:objects}
\end{table*}

\begin{table*}
\vspace{15pt}
\centering
\caption{Common hyperparameters}
\begin{tabular}{c c}
\hline
\textbf{Hyperparameter} & \textbf{Value}\\
\hline
& \\
GMM number of modes & $5$\\
Image encoder & ResNet-18\\
Random crop ratio & $90$\% of image height \\
& \\
Optimizer & Adam\\
Batch size & $16$\\

\# Training steps per epoch & $500$\\
\# Total training epochs & $1000$ \\
& \\
Evaluation checkpoint interval (in epoch) & $50$ \\
& \\

\hline
\end{tabular}
\label{table:common}
\end{table*}

\begin{table*}
\vspace{15pt}
\centering
\caption{BC backbone hyperparameters}
\begin{tabular}{c c}
\hline
\textbf{Hyperparameter} & \textbf{Value}\\
\hline
& \\
RNN hidden dim & $1000$ \\
RNN sequence length & $10$\\
\# of LSTM layers & $2$ \\
Learning rate & $1\mathrm{e}{-4}$\\
& \\

\hline
\end{tabular}
\label{table:bc-hyper}
\end{table*}

\begin{table*}
\vspace{15pt}
\centering
\caption{IQL hyperparameters}
\begin{tabular}{c c}
\hline
\textbf{Hyperparameter} & \textbf{Value}\\
\hline
& \\
Reward scale & $1.0$ \\
Termination & false \\
Discount factor $r$ & $0.99$ \\
Beta $\beta$ & $1.0$ \\
Adv filter & exponential \\
V function quantile & $0.75$ \\
& \\
Actor lr & $1\mathrm{e}{-4}$\\
Actor lr decay factor & $0.1$\\
Actor mlp layers & $[1024, 1024]$ \\
& \\
Critic lr & $1\mathrm{e}{-4}$\\
Critic lr decay factor & $0.1$\\
Critic mlp layers & $[1024, 1024]$ \\
& \\

\hline
\end{tabular}
\label{table:iql}
\end{table*}

\clearpage

\begin{table*}
\vspace{15pt}
\centering
\caption{Task hyperparameters}
\begin{tabular}{c c c c c}
\hline
\textbf{Hyperparameter} & \textbf{Nut Assembly} & \textbf{ToolHang} & \textbf{Gear Insertion (Real)} & \textbf{Coffee Pod Packing (Real)}\\
\hline
& \\
Image size ($h \times w$) & $84 \times 84 $ & $128 \times 128$ & $84 \times 84$ & $84 \times 84$ \\
Initial \# of human demonstrations & $50$ & $80$ & $30$ & $30$ \\
Evaluation rollout length & $400$ & $700$ & $1000$ & $1000$ \\
& \\

\hline
\end{tabular}
\label{table:task-dependent}
\end{table*}


\end{document}